\documentclass{article} 
\usepackage{iclr2020_conference,times}
\usepackage{graphicx}
\usepackage{subcaption}

\usepackage{amsmath,amsfonts,bm}

\def\eqref#1{equation~\ref{#1}}

\def\1{\bm{1}}

\DeclareMathAlphabet{\mathsfit}{\encodingdefault}{\sfdefault}{m}{sl}
\SetMathAlphabet{\mathsfit}{bold}{\encodingdefault}{\sfdefault}{bx}{n}

\usepackage{longtable}
\usepackage{hyperref}
\usepackage{url}

\title{Building Disaster Damage Assessment in Satellite Imagery with Multi-Temporal Fusion}

\author{Ethan Weber \\
Massachusetts Institue of Technology \\
\texttt{ejweber@mit.edu} \\
\And
Hassan Kané \\
WL Research \\
\texttt{hassanmohamed@alum.mit.edu} \\
}

\iclrfinalcopy 
\begin{document}

\maketitle

\begin{abstract}
Automatic change detection and disaster damage assessment are currently procedures requiring a huge amount of labor and manual work by satellite imagery analysts. In the occurrences of natural disasters, timely change detection can save lives. In this work, we report findings on problem framing, data processing and training procedures which are specifically helpful for the task of building damage assessment using the newly released xBD dataset. Our insights lead to substantial improvement over the xBD baseline models, and we score among top results on the xView2 challenge leaderboard. We release our code used for the competition\footnote{\url{https://github.com/ethanweber/xview2}}.
\end{abstract}

\section{Introduction}

When a natural disaster occurs, quick and accurate information is critical to an effective response. To better deploy resources in affected areas, it is important for emergency responders to know the location and severity of the damages.

Satellite imagery offers a powerful source of information and can be used to assess the extent and areas of damages. However, a common bottleneck in the current workflows involves the time it takes for human analysts to observe an affected area and identify damaged zones. This process can take hours in a situation where time is of the essence. It therefore presents room to be accelerated through leveraging artificial intelligence.

In the last several years, Convolutional Neural Networks \citep{krizhevsky2012imagenet} have achieved human level performance on a variety of computer vision tasks, including object recognition and image segmentation \citep{lecun2015deep}. These techniques are highly relevant and applicable in the case of satellite image analysis for disaster damage assessment (\cite{ji2018identifying}; \cite{cooner2016detection}).

The main contributions of this work is a characterization of the importance of mono-temporal vs. multi-temporal settings when performing damage assessment with natural disaster along with insights on helpful image pre-processing techniques. The specific insights in this work are that substantially better performance is obtained by independently feeding the pre and post-disaster images through a CNN with shared weights. The features are then fused before a semantic segmentation final layer. Furthermore, working on smaller image crops and weighting error on damage classes inversely proportional to their statistical occurrence in the training dataset leads to models strongly improving over baseline models.

\begin{figure*}[t]
\centering
\includegraphics[width=\textwidth]{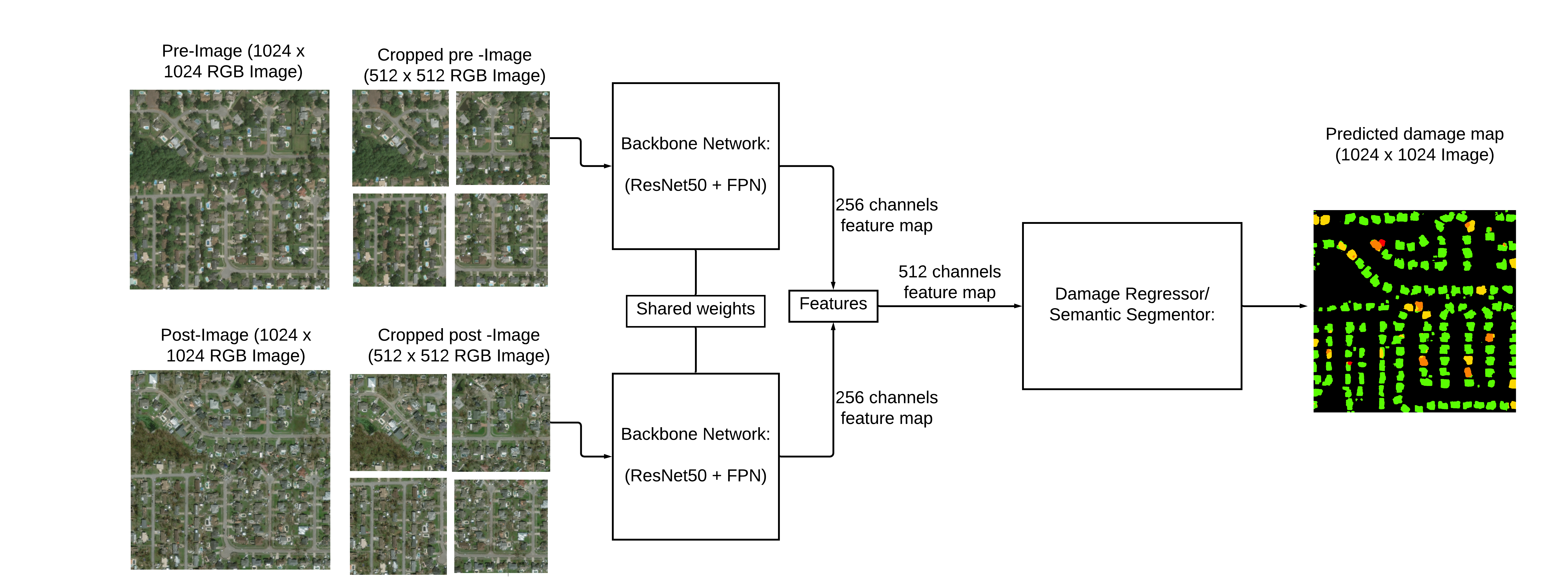}
\caption{\textbf{Our model architecture.} For building damage assessment, we feed pre and post-disaster images through a ResNet50 backbone with shared weights. We then concatenate features before semantic segmentation to obtain our final building damage predictions. The output is a 5-class damage map ranging from no building (0) to destroyed building (4).}
\label{fig:our_model}
\end{figure*}

\section{Related Work}

In this section, we highlight related work in computer vision and large-scale datasets for disaster damage analysis. In particular, we focus on building disaster damage assessment.

\subsection{Computer Vision Techniques for Building Damage Assessment}

Researchers have applied machine learning approaches to building damage assessment in satellite
imagery. \cite{xu2019building} described a method to build convolutional neural networks that automatically detect damaged buildings in satellite images using pre and post images and showed that the model can generalize well to new regions and disasters if it is fine-tuned on a small set of examples from that region. Their work is done on a proprietary dataset spanning three disasters and is framed as a binary pixel classification over damaged/not damaged buildings.

\cite{cooner2016detection} compared the performance of multiple machine learning methods in building damage detection with both pre and post-event satellite imagery of the 2010 Haiti earthquake, and found that a feed-forward neural network achieved the lowest error rate of 40\%. \cite{ji2018identifying} developed a convolutional network to identify collapsed buildings from post-event satellite imagery of the Haiti earthquake, and reached an overall accuracy of 78.6\%.  \cite{duarte2018satellite} combine drone and satellite images of disasters to improve the accuracy of their convolutional networks, with a best reported accuracy of 94.4\%. In these settings damaged building detection over images is framed as a binary pixel classification problem.

\cite{yang2018multi} show that convolutional neural network (CNN) can perform feature-based multi-temporal remote sensing image registration and can outperform four state-of-the-art methods in most scenarios. \cite{nia2017building} show that convolutional neural networks can perform damage assessment on post disaster image. However, their dataset uses ground-level images instead of satellite images.

\subsection{Datasets for damage assessment}
Until xBD \citep{gupta2019xbd}, adequate satellite imagery addressing building damage was not generally available. Other satellite imagery datasets were often limited to single disaster types and did not have common criteria for assessing damage assessment (\cite{fujita2017damage}; \cite{chen2018benchmark}; \cite{foulser2012use}). This made comparing model and agreeing on a problem framing difficult.

The xview dataset \citep{lam2018xview} is a precursor to the xBD dataset. It is one of the largest and most diverse publicly available object detection datasets, with over 1 million objects across 60 classes in over 1,400 km$^2$ of imagery. It focuses on object detection and not on building damage assessment.

The complete xBD dataset contains satellite images from 19 different natural disasters across 22,068 images and contains 850,736 building polygons. Each image has a 1024 by 1024 pixels resolution. Images are collected from WorldView-3 satellites at 0.3m spatial resolution. The imagery covers a total of 45,361.79 km$^2$. It also introduces a four-level damage annotation scale.

\section{Model}

In this section, we explain the task of building damage assessment and data pre-processing. Given the nature of the xBD dataset, with pre-disaster and post-disaster images, we exploit the multi-temporal information to predict both building locations and damage level with the same network. For localization the task is to predict 0 or 1 (background or building) and for damage assessment the task is to predict a 5 dimensional output: 0 is no building and 1-4 is the damage level (1: undamaged, 2: minor, 3: major: 4: destroyed). An alternative approach, which we found to be the best, is focus on only damage assessment but say damage levels 1-4 are ``buildings". This avoids the problem of using completely separate networks, as done in the xBD baselines \citep{gupta2019xbd}. We find this very important to our final results.

\begin{table*}[ht]
\resizebox{\textwidth}{!}{%
\begin{tabular}{| c c c c c | c c c |}
\hline
Architectures & Two-image input (pre and post) & 512 x 512 images & Joint prediction & Class-specific weighting & Overall F1 & Localization F1 & Damage F1 \\ \hline
Instance seg. &                             &                 &                 &                         & 0.492      & 0.705           & 0.401     \\
Semantic seg. &                             &                 &                 &                         & 0.536      & 0.819           & 0.414     \\
Semantic seg. &                             & \checkmark      &                 &                         &            & 0.835           &           \\
Semantic seg. & \checkmark                  & \checkmark      & \checkmark                &                          & 0.729      & 0.847           & 0.679     \\
Semantic seg. & \checkmark                  & \checkmark                & \checkmark                & \checkmark                        & 0.738      & 0.835           & 0.697   \\ \hline
\end{tabular}%
}
\caption{\textbf{Our ablation study.} In this table, we show the experiments that inform our final model. The best model uses semantic segmentation, both pre and post image as input, 512 x 512 image crops, joint building localization and damage prediction, and cross-entropy loss weightings to handle class-imbalance.}\label{tab:ablation_study}
\end{table*}

\subsection{Baseline Architecture}

The first decision in this work is related to problem framing. For this work, we are focused on building damage assessment, which means we care about both (i) building localization and (ii) building per-pixel damage classification. Because we have two tasks (i.e. identifying the building and determining the damage level), we decided to use the same model architecture for both tasks. However, the baseline model uses two separate networks.

The task and baseline models are described in the xBD paper \citep{gupta2019xbd}. For localization, a U-Net \citep{ronneberger2015unet} architecture is used for binary pixel classification of ``background" or ``building". For damage assessment, the model is less straightforward. It uses post-damage images fed into a ResNet-50 backbone \citep{he2016deep} pre-trained on ImageNet \citep{deng2009imagenet} and additional features from a shallow CNN. All convolutional layers use a ReLU activation. The output is a one-hot encoded vector where each element represents the probability of an ordinal class. The model uses an ordinal cross-entropy loss function. Unlike traditional cross-entropy, ordinal cross-entropy penalizes relative to the distance between true and predicted ordinal class. Since the difference between any two classes is not interchangeable, this loss function allows the model to distinguish between the different levels of damage.

Extending beyond this baseline, we decided that using the same network for both building detection and damage assessment was a more natural formulation of the problem. This way the model can jointly reason about similar features. For this reason, we use a Mask R-CNN backbone \citep{he2017mask} augmented with a Feature Pyramid Network module \citep{lin2017feature} and a semantic segmentation head. The model was pretrained on the ImageNet dataset \citep{deng2009imagenet}. The implementation of this Mask R-CNN architecture was obtained using the Detectron2 \citep{wu2019detectron2} library from Facebook AI Research which has PyTorch \citep{paszke2019pytorch} implementations of other models as well. With this backbone, we tried both instance segmentation and semantic segmentation to generate the per-pixel classification output. Our final model uses semantic segmentation, as it's a more natural damage-assessment formulation without the notion of instances. Often time, the buildings are too small for instance segmentation to be appropriate.

For the loss function, we use cross-entropy loss on the predicted classes with the ground truth labels.

\subsection{Data Processing Technique}

With the xBD dataset, we have both pre-disaster and post-disaster images available, which can be used for the two tasks of building localization and damage assessment. The images are 1024 by 1024, but we found that the buildings were often too small resolution for the model to accurately draw building boundaries. For this reason, we trained and ran models on 4 512 by 512 images forming the top-left, top-right, bottom-left and bottom-right quadrants.

Furthermore, given this pre and post data, damage assessment can be framed as either a mono-temporal and multi-temporal task \citep{xu2019building}. In the mono-temporal setting, only the post images are fed to a model that has to predict damage level per pixel. In the multi-temporal setting, both the pre and post images are fed to the model that has to predict damage levels on the post images. In this case we fed both the pre and post images independently through the R-CNN base, with shared weights. These features were then concatenated before being fed through the semantic segmentation head of the networks. We show our best architecture in Figure \ref{fig:our_model}.

\begin{table*}[ht]
\resizebox{\textwidth}{!}{%
\begin{tabular}{|l|l|l|l|l|l|l|}
\hline
& F-1 score (overall) & No Damage F1 & Minor Damage F1 & Major Dmg F1 & Destroyed F1 \\  \hline
xBD Baseline  &         0.265  &         0.663  &         0.144  &         0.009  &         0.466 \\   \hline
Our model &         \textbf{0.741}  &         \textbf{0.906}  &         \textbf{0.493}  &        \textbf{0.722}  &         \textbf{0.837} \\
\hline
\end{tabular}%
}
\caption{\textbf{Comparison with xBD baseline.} On all metrics, our best model outperforms the xBD baseline model by a large margin.}\label{tab:correlations}
\end{table*}

\section{Experiments}
In this section, we explain the dataset and experiments used to inform the final model design. The metrics used are for building localization and damage assessment from the xView2 competition\footnote{\url{https://xview2.org/}} (\citep{gupta2019xbd}).

\subsection{xBD Dataset}
This work is one of the first to apply recent deep learning techniques to the xBD dataset. xBD is the largest and first building damage assessment dataset released to date. It contains 1024 x 1024 satellite imagery with 30cm per pixel resolution covering a diverse set of disasters--including earthquakes, floods, volcanic eruption, wildfire and wind--and geographical locations (16) with over 850,000 building annotations (on a 1 to 4 damage scale) across over 45,000 km$^2$  of imagery before and after disasters.

\subsection{Ablation Study}
For the task of multi-class pixel classification, both instance segmentation and semantic segmentation are common solutions. The goal of instance segmentation is assign a class and instance ID to every pixel. Similarly, the goal of semantic segmentation is to assign a class to every pixel, but in this case the notion of instance ID is not used. The nature of damage assessment is more suited for semantic segmentation, where instances IDs are not relevant.

Given the success of instance segmentation networks such as Mask R-CNN \citep{he2017mask}, we perform experiments with both instance segmentation and semantic segmentation architectures. Furthermore, we experiment with using image crops, both pre and post images as the input, predicting localization and damage together, and weighting the class loss inversely proportional to the training set distribution.

Looking at Table \ref{tab:ablation_study}, we see the different experiments used to choose our best model. Notice that the F1 score is computed with the xBD test set withheld for the xView2 competition. Localization F1 score is for building detection (0-1) and damage F1 is for damage assessment (0-4). When using a joint prediction, we classify all pixels with damage level at least 1, to be a building. This is how we use the same network for both tasks. Overall F1 is a weighted combination of 30\% localization F1 and 70\% damage F1.

In the first row (row 1), instance segmentation is used on full 1024 x 1024 images. In the case of overlapping bounding boxes, we use the label with higher damage prediction. By switching to semantic segmentation (row 2), we notice a large improvement in localization (0.114). This is due to the building box predictions suffering on small building sizes. By using 4 512 x 512 crops per image (row 3), we obtain a boost in localization F1 of 0.016. By using our multi-temporal input with pre and post images and adding the joint prediction (row 4), we see an increase for all metrics. Lastly (row 5), we add class-specific cross-entropy weighting to obtain an increase in damage and overall F1, while only losing slight performance in localization F1. This is because we put more weighting on building damage classification. The reason for this weighting is that the 73.6\% of polygons have no damage. Furthermore, most of the image is the ``no building" class.

We don't show this in the table, but simple concatenation of pre and post images to produce a 6-channel input produced very poor results. Similarly, subtracting the pre and post images before input did not work well either. Our best results were obtained when processing pre and post images individually, and then concatenating the features before segmentation. Instead of a single-network 256 channel feature embedding being used with the semantic segmentation head, we use 512 features from stacking the pre and post image features coming from the FPN before the semantic segmentation head, which predicts the final 5 classes.

\subsection{Training}

We performed most experiments on a machine with 4 NVIDIA 1080 TIs, and training takes roughly 6 hours to convergence when using ImageNet pretrained weights. When training for too long, the network will collapse into predicting all 0 (no building) labels. Our inversely proportional class weightings allows the network to train longer before collapsing, but it doesn't solve the issue entirely. Code will be released for replication of our work.

\section{Results}

With our final model, we obtain a localization F1 of 0.835 and damage F1 of 0.697. These metrics are combined to obtain an overall F1 of 0.738 in the xView2 competition with the xBD test set. In Table \ref{tab:correlations}, we show the breakdown of the building damage assessment result compared to the xBD baseline. The metrics are computed on the holdout set, and we outperform baselines by a large margin. Notice that this is compared to the xBD baseline, but our results are competitive on the xView2 leaderboard, in which we were ranked place 2 in Track 3: "Evaluation Only", and 40th place before non-validated submissions were removed. Our code is available at \url{https://github.com/ethanweber/xview2}.

\section{Conclusion}
In conclusion, the four main insights behind our performance include working with image quadrants instead of the full image, using one architecture trained on both the pre and post images and fused before the final segmentation layer, using the Mask R-CNN with FPN architecture and engineering our loss function to weight errors on classes inversely proportional to their occurrence on the dataset. Future research directions include using an ordinal cross-entropy loss function to penalize errors in the damage scales differently and also experimenting with other ways to combine the information from the pre and post images at different stages of feature extraction. Incorporating the disaster type (e.g. flood, fire, etc.) into the building damage assessment prediction could also be an interesting direction.

\bibliography{iclr2020_conference}

\begin{thebibliography}{20}
\providecommand{\natexlab}[1]{#1}
\providecommand{\url}[1]{\texttt{#1}}
\expandafter\ifx\csname urlstyle\endcsname\relax
  \providecommand{\doi}[1]{doi: #1}\else
  \providecommand{\doi}{doi: \begingroup \urlstyle{rm}\Url}\fi

\bibitem[Chen et~al.(2018)Chen, Escay, Haberland, Schneider, Staneva, and
  Choe]{chen2018benchmark}
Sean~Andrew Chen, Andrew Escay, Christopher Haberland, Tessa Schneider,
  Valentina Staneva, and Youngjun Choe.
\newblock Benchmark dataset for automatic damaged building detection from
  post-hurricane remotely sensed imagery.
\newblock \emph{arXiv preprint arXiv:1812.05581}, 2018.

\bibitem[Cooner et~al.(2016)Cooner, Shao, and Campbell]{cooner2016detection}
Austin~J Cooner, Yang Shao, and James~B Campbell.
\newblock Detection of urban damage using remote sensing and machine learning
  algorithms: Revisiting the 2010 haiti earthquake.
\newblock \emph{Remote Sensing}, 8\penalty0 (10):\penalty0 868, 2016.

\bibitem[Deng et~al.(2009)Deng, Dong, Socher, Li, Li, and
  Fei-Fei]{deng2009imagenet}
Jia Deng, Wei Dong, Richard Socher, Li-Jia Li, Kai Li, and Li~Fei-Fei.
\newblock Imagenet: A large-scale hierarchical image database.
\newblock In \emph{2009 IEEE conference on computer vision and pattern
  recognition}, pp.\  248--255. Ieee, 2009.

\bibitem[Duarte et~al.(2018)Duarte, Nex, Kerle, and
  Vosselman]{duarte2018satellite}
D~Duarte, Francesco Nex, N~Kerle, and George Vosselman.
\newblock Satellite image classification of building damages using airborne and
  satellite image samples in a deep learning approach.
\newblock \emph{ISPRS Annals of Photogrammetry, Remote Sensing \& Spatial
  Information Sciences}, 4\penalty0 (2), 2018.

\bibitem[Foulser-Piggott et~al.(2012)Foulser-Piggott, Spence, Saito, Brown, and
  Eguchi]{foulser2012use}
R~Foulser-Piggott, R~Spence, K~Saito, DM~Brown, and R~Eguchi.
\newblock The use of remote sensing for post-earthquake damage assessment:
  lessons from recent events, and future prospects.
\newblock In \emph{Proceedings of the Fifthteenth World Conference on
  Earthquake Engineering}, pp.\ ~10, 2012.

\bibitem[Fujita et~al.(2017)Fujita, Sakurada, Imaizumi, Ito, Hikosaka, and
  Nakamura]{fujita2017damage}
Aito Fujita, Ken Sakurada, Tomoyuki Imaizumi, Riho Ito, Shuhei Hikosaka, and
  Ryosuke Nakamura.
\newblock Damage detection from aerial images via convolutional neural
  networks.
\newblock In \emph{2017 Fifteenth IAPR International Conference on Machine
  Vision Applications (MVA)}, pp.\  5--8. IEEE, 2017.

\bibitem[Gupta et~al.(2019)Gupta, Hosfelt, Sajeev, Patel, Goodman, Doshi, Heim,
  Choset, and Gaston]{gupta2019xbd}
Ritwik Gupta, Richard Hosfelt, Sandra Sajeev, Nirav Patel, Bryce Goodman, Jigar
  Doshi, Eric Heim, Howie Choset, and Matthew Gaston.
\newblock xbd: A dataset for assessing building damage from satellite imagery.
\newblock \emph{arXiv preprint arXiv:1911.09296}, 2019.

\bibitem[He et~al.(2016)He, Zhang, Ren, and Sun]{he2016deep}
Kaiming He, Xiangyu Zhang, Shaoqing Ren, and Jian Sun.
\newblock Deep residual learning for image recognition.
\newblock In \emph{Proceedings of the IEEE conference on computer vision and
  pattern recognition}, pp.\  770--778, 2016.

\bibitem[He et~al.(2017)He, Gkioxari, Doll{\'a}r, and Girshick]{he2017mask}
Kaiming He, Georgia Gkioxari, Piotr Doll{\'a}r, and Ross Girshick.
\newblock Mask r-cnn.
\newblock In \emph{Proceedings of the IEEE international conference on computer
  vision}, pp.\  2961--2969, 2017.

\bibitem[Ji et~al.(2018)Ji, Liu, and Buchroithner]{ji2018identifying}
Min Ji, Lanfa Liu, and Manfred Buchroithner.
\newblock Identifying collapsed buildings using post-earthquake satellite
  imagery and convolutional neural networks: A case study of the 2010 haiti
  earthquake.
\newblock \emph{Remote Sensing}, 10\penalty0 (11):\penalty0 1689, 2018.

\bibitem[Krizhevsky et~al.(2012)Krizhevsky, Sutskever, and
  Hinton]{krizhevsky2012imagenet}
Alex Krizhevsky, Ilya Sutskever, and Geoffrey~E Hinton.
\newblock Imagenet classification with deep convolutional neural networks.
\newblock In \emph{Advances in neural information processing systems}, pp.\
  1097--1105, 2012.

\bibitem[Lam et~al.(2018)Lam, Kuzma, McGee, Dooley, Laielli, Klaric, Bulatov,
  and McCord]{lam2018xview}
Darius Lam, Richard Kuzma, Kevin McGee, Samuel Dooley, Michael Laielli, Matthew
  Klaric, Yaroslav Bulatov, and Brendan McCord.
\newblock xview: Objects in context in overhead imagery.
\newblock \emph{arXiv preprint arXiv:1802.07856}, 2018.

\bibitem[LeCun et~al.(2015)LeCun, Bengio, and Hinton]{lecun2015deep}
Yann LeCun, Yoshua Bengio, and Geoffrey Hinton.
\newblock Deep learning.
\newblock \emph{nature}, 521\penalty0 (7553):\penalty0 436--444, 2015.

\bibitem[Lin et~al.(2017)Lin, Doll{\'a}r, Girshick, He, Hariharan, and
  Belongie]{lin2017feature}
Tsung-Yi Lin, Piotr Doll{\'a}r, Ross Girshick, Kaiming He, Bharath Hariharan,
  and Serge Belongie.
\newblock Feature pyramid networks for object detection.
\newblock In \emph{Proceedings of the IEEE conference on computer vision and
  pattern recognition}, pp.\  2117--2125, 2017.

\bibitem[Nia \& Mori(2017)Nia and Mori]{nia2017building}
Karoon~Rashedi Nia and Greg Mori.
\newblock Building damage assessment using deep learning and ground-level image
  data.
\newblock In \emph{2017 14th Conference on Computer and Robot Vision (CRV)},
  pp.\  95--102. IEEE, 2017.

\bibitem[Paszke et~al.(2019)Paszke, Gross, Massa, Lerer, Bradbury, Chanan,
  Killeen, Lin, Gimelshein, Antiga, et~al.]{paszke2019pytorch}
Adam Paszke, Sam Gross, Francisco Massa, Adam Lerer, James Bradbury, Gregory
  Chanan, Trevor Killeen, Zeming Lin, Natalia Gimelshein, Luca Antiga, et~al.
\newblock Pytorch: An imperative style, high-performance deep learning library.
\newblock In \emph{Advances in Neural Information Processing Systems}, pp.\
  8024--8035, 2019.

\bibitem[Ronneberger et~al.(2015)Ronneberger, Fischer, and
  Brox]{ronneberger2015unet}
Olaf Ronneberger, Philipp Fischer, and Thomas Brox.
\newblock U-net: Convolutional networks for biomedical image segmentation.
\newblock \emph{CoRR}, abs/1505.04597, 2015.
\newblock URL \url{http://arxiv.org/abs/1505.04597}.

\bibitem[Wu et~al.(2019)Wu, Kirillov, Massa, Lo, and
  Girshick]{wu2019detectron2}
Yuxin Wu, Alexander Kirillov, Francisco Massa, Wan-Yen Lo, and Ross Girshick.
\newblock Detectron2.
\newblock \url{https://github.com/facebookresearch/detectron2}, 2019.

\bibitem[Xu et~al.(2019)Xu, Lu, Li, Khaitan, and Zaytseva]{xu2019building}
Joseph~Z Xu, Wenhan Lu, Zebo Li, Pranav Khaitan, and Valeriya Zaytseva.
\newblock Building damage detection in satellite imagery using convolutional
  neural networks.
\newblock \emph{arXiv preprint arXiv:1910.06444}, 2019.

\bibitem[Yang et~al.(2018)Yang, Dan, and Yang]{yang2018multi}
Zhuoqian Yang, Tingting Dan, and Yang Yang.
\newblock Multi-temporal remote sensing image registration using deep
  convolutional features.
\newblock \emph{IEEE Access}, 6:\penalty0 38544--38555, 2018.

\end{thebibliography}
\bibliographystyle{iclr2020_conference}
\end{document}